\documentclass[10pt,twocolumn,letterpaper]{article}
\usepackage{iccv}
\usepackage{booktabs}
\usepackage{times}
\usepackage{epsfig}
\usepackage{graphicx}
\usepackage{amsmath}
\usepackage{amssymb}
\usepackage{fontawesome5}
\usepackage{caption}
\usepackage{eso-pic}
\usepackage[normalem]{ulem}
\useunder{\uline}{\ul}{}
\usepackage{wrapfig}
\BeforeBeginEnvironment{wraptable}{\setlength{\intextsep}{0pt}}
\BeforeBeginEnvironment{wrapfigure}{\setlength{\intextsep}{0pt}}
\usepackage{blindtext}
\usepackage{multirow}
\usepackage{bm}
\usepackage{bbm} 
\usepackage{comment}
\usepackage{algorithm}
\usepackage{algorithmic}
\usepackage{subcaption}
\usepackage{array} 
\usepackage[dvipsnames]{xcolor}
\usepackage{color, colortbl}
\usepackage[misc,geometry]{ifsym} 

\makeatletter
\renewcommand\paragraph{
  \@startsection{paragraph} 
  {4} 
  {\z@} 
  {.5em \@plus1ex \@minus.2ex} 
  {-1.5em} 
  {\normalfont\normalsize\bfseries} 
}
\makeatother

\makeatletter
\def\@fnsymbol#1{\ensuremath{\ifcase#1\or \textsuperscript{~\Letter}\or \ddagger\or
   \mathsection\or \mathparagraph\or \|\or **\or \dagger\dagger
   \or \ddagger\ddagger \else\@ctrerr\fi}}
\makeatother

\newcommand{\tableCellHeight}{1}
\newcommand{\tabstyle}[1]{
  \setlength{\tabcolsep}{#1}
  \renewcommand{\arraystretch}{\tableCellHeight}
  \centering
  \small
}

\definecolor{tabhighlight}{HTML}{e5e5e5}
\definecolor{citecolor}{HTML}{0071bc}

\usepackage[pagebackref=true,breaklinks=true,letterpaper=true,colorlinks,bookmarks=false]{hyperref}

\iccvfinalcopy 

\def\iccvPaperID{7480} 

\ificcvfinal\fi

\begin{document}
\title{Knowledge-Aware Prompt Tuning for Generalizable Vision-Language Models}

\author{Baoshuo Kan$^{1*}$,\ \ Teng Wang$^{2,3*}$,\ \ Wenpeng Lu$^{1\dagger}$,\ \ Xiantong Zhen$^{4}$, Weili Guan$^{5}$,\ \ Feng Zheng$^{2\dagger}$ \\
$^{1}$Qilu University of Technology (Shandong Academy of Sciences) \ \  \\
$^{2}$Southern University of Science and Technology\ \ 
$^{3}$The University of Hong Kong \\
$^{4}$United Imaging Healthcare 
$^{5}$Monash Univerisity \\
\tt\small 10431200583@stu.qlu.edu.cn\ \ tengwang@connect.hku.hk\ \ wenpeng.lu@qlu.edu.cn \\
\tt\small zhenxt@gmail.com \ \ weili.guan@monash.edu \ \ f.zheng@ieee.org}

\maketitle
\ificcvfinal\thispagestyle{empty}\fi

\begin{abstract}

\let\thefootnote\relax\footnotetext{$^*$ Equal contribution. $^{\dagger}$ Corresponding author. Work done when Baoshuo Kan visited to Feng Zheng Lab in SUSTech.} 

   Pre-trained vision-language models, e.g., CLIP, working with manually designed prompts have demonstrated great capacity of transfer learning. Recently, learnable prompts achieve state-of-the-art performance, which however are prone to overfit to seen classes, failing to generalize to unseen classes. In this paper, we propose a Knowledge-Aware Prompt Tuning (KAPT) framework for vision-language models. Our approach takes the inspiration from human intelligence in which external knowledge is usually incorporated into recognizing novel categories of objects. Specifically, we design two complementary types of knowledge-aware prompts for the text encoder to leverage the distinctive characteristics of category-related external knowledge. The discrete prompt extracts the key information from descriptions of an object category, and the learned continuous prompt captures overall contexts. We further design an adaptation head for the visual encoder to aggregate salient attentive visual cues, which establishes discriminative and task-aware visual representations. We conduct extensive experiments on 11 widely-used benchmark datasets and the results verify the effectiveness in few-shot image classification, especially in generalizing to unseen categories. Compared with the state-of-the-art CoCoOp method, KAPT exhibits favorable performance and achieves an absolute gain of 3.22\% on new classes and 2.57\% in terms of harmonic mean.
\end{abstract}

\section{Introduction}
Recently, large-scale pre-trained vision-language models, \eg, CLIP \cite{radford2021learning}, ALIGN \cite{jia2021scaling}, and FLIP \cite{yao2021filip}, have demonstrated remarkable performance in zero/few-shot learning tasks. Unlike traditional vision-only frameworks that are trained mainly by a closed set of single-modal data, vision-language models train two uni-modal encoders on massive amounts of image-text pairs to exploit cross-modal alignments in the semantic space. By leveraging large-scale web-scale image-text data, pre-trained vision-language models are endowed with the ability to solve zero/few-shot downstream tasks and even recognize open-set visual concepts \cite{radford2021learning, jia2021scaling, yao2021filip}. 
Expressly, when a new classification task arrives, the CLIP text-encoder encodes manually designed textual prompt (\eg, ``\textit{a photo of a [\textit{label}]}."), and then cosine similarity between textual features and image features is computed. However, identifying appropriate manually designed prompts is an art that requires both domain expertise and laborious prompt engineering.

\begin{figure}[]

	\centering
	\includegraphics[width=0.99\linewidth]{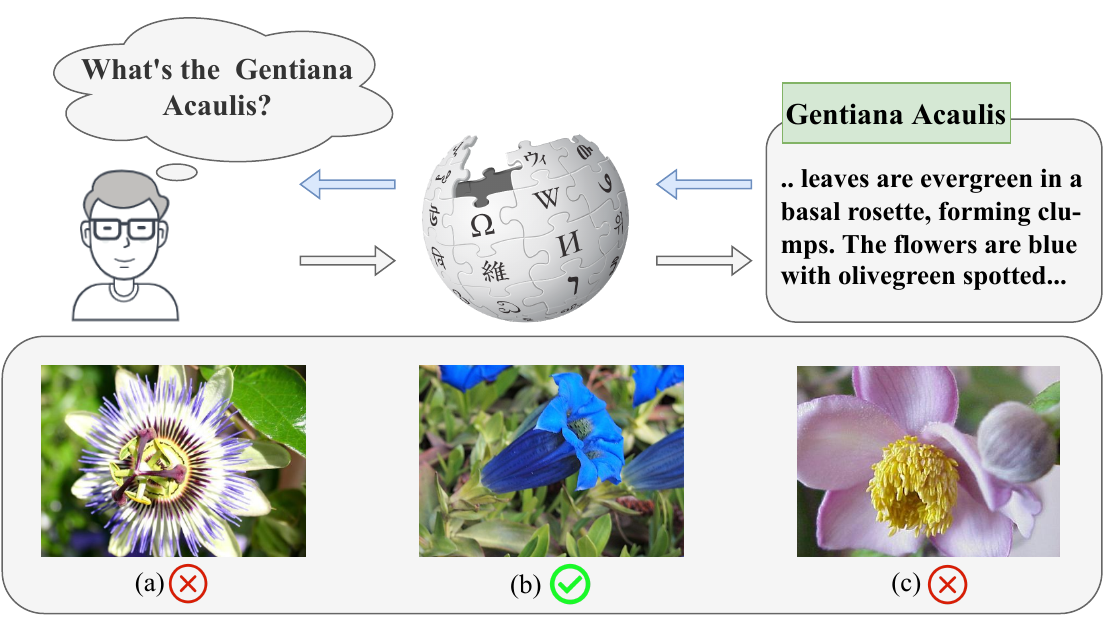}
	\caption{\textbf{A motivating example.} The textual description of the ``Gentiana Acaulis'' conduces to the recognition of the corresponding image (b) of  Gentiana Acaulis.} \label{fig1}
\vspace{-0.5cm}
\end{figure}

To avoid the manual prompt design, some recent research (\eg, CoOp \cite{zhou2022learning}) on visual representations are mainly inspired by prompt tuning approaches \cite{zhong2021factual, lester2021power, li2021prefix} in Natural Language Processing (NLP), like learnable prompts. By optimizing their models with learnable prompts in closed datasets, these methods achieve outstanding performance in seen classes. However, the learned prompts are usually prone to overfit to the seen classes and suffer from insufficient generalization ability to unseen classes under the same task. 

Recently, CoCoOp \cite{zhou2022conditional} was developed to improve the generalizability. The model constructs specific prompts by conditioning them on each instance, which achieves stronger robustness to category shift. Nonetheless, their learnable prompts are shared across all categories, which leads to weak discrimination between distinct characteristics of different categories; meanwhile, the model is incapable of perceiving factual details for class labels due to lack of fine-grained textual information on category, especially for uncommon classes that are rarely encountered during pre-training or have poor relevance to seen classes. Thus, the CoCoOp model still falls short of the generalizability to unseen scenarios.

\begin{figure}[]

	\centering
	\includegraphics[width=0.99\linewidth]{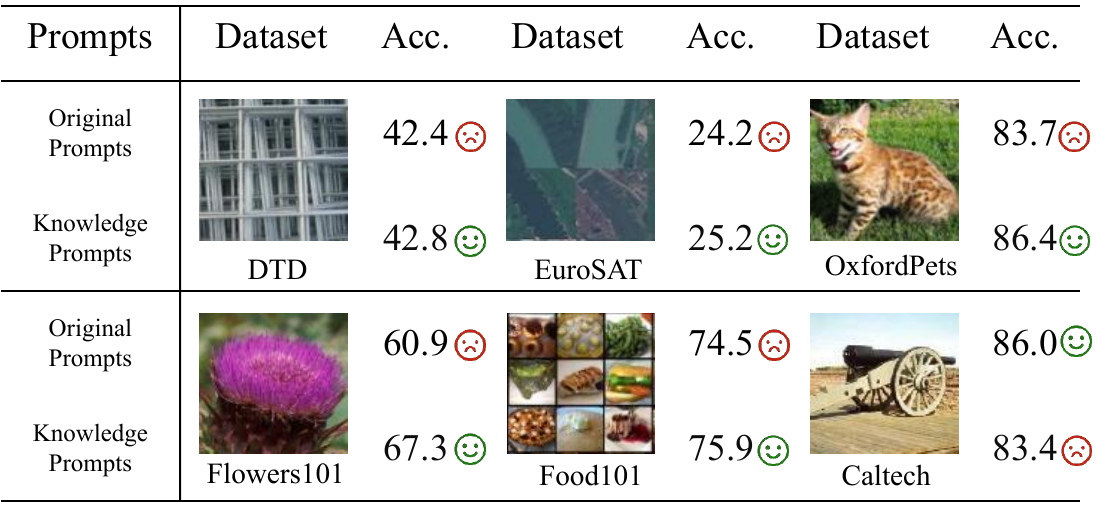}
	\caption{\textbf{External knowledge improves the generalizability of CLIP.} We perform zero-shot classification with discrete prompts on six image classification datasets. The original prompt from CLIP~\cite{radford2021learning} is \emph{a photo of a [label]}. \textcolor{black}{The proposed knowledge prompt concatenates category-related external knowledge from the Wikipedia Encyclopedia with the original prompt.}}
	 \label{fig2}
\vspace{-0.5cm}
\end{figure}

Inspired by how humans utilize knowledge bases to learn novel visual concepts, we propose to incorporate external knowledge into prompt learning by leveraging accurate descriptions of concepts and their contextual relationships to overcome the aforementioned issues. As a motivating example, we can see in Figure~\ref{fig1} that it is generally hard for humans to imagine the visual appearance of uncommon categories, and even harder to recognize them when seeing a scientific name for the first time (\eg, Gentiana acaulis). However, once we learn the key characteristics by reading the textual description from the knowledge base, it becomes much easier to recognize the images and the correspondence between different categories. { Considering that textual descriptions of scientific names are unparalleled and contain identifying authentication information, we assign unique knowledge for each category to improve the generalization ability by enhancing the discrimination of prompts.}

Specifically, we present a novel, Knowledge-Aware Prompt Tuning (KAPT) framework for vision-language models. We first retrieve encyclopedic knowledge related to category names from Wikipedia Encyclopedia containing a large number of entity descriptions. To take full advantage of category-related external knowledge, {we design two complementary types of prompts: discrete and learnable continuous prompts. Discrete prompts carry the summarized texts that directly describe the visual appearance of the category}, and learnable continuous prompts carry contextual information that may cover a broader background of the category. As shown in Figure~\ref{fig2}, a preliminary experiment verifies that the proposed discrete knowledge-aware prompt improves performance of CLIP on several image datasets. Meanwhile, to further adapt the visual representation towards a specific task {for inhibiting disturbance of task-irrelevant visual concepts,} we propose an adaptation head that refines the image features by attending to the salient visual cues relevant to categories of the target task.

The main contributions can be summarized as follows:
\begin{itemize}

    \item [$\bullet$] We propose a novel prompt tuning framework for vision-language models by incorporating external knowledge, which greatly improves the generalizability on unseen object categories.
    
    \item [$\bullet$] We design two complementary types of knowledge-aware prompts, which enables the model to fully exploit category-related dense knowledge retrieved from  the Wikipedia Encyclopedia.

    \item[$\bullet$] We further propose a task-aware visual adaptation head to aggregate the attentive visual features conditioned on linguistic description of categories, which maximally capture task-related visual cues, while suppressing the disturbance caused by task-irrelevant visual concepts.
	
    \item [$\bullet$] Extensive experimental results on 11 popular image datasets demonstrate the effectiveness of the proposed method. Our method significantly outperforms state-of-the-art methods on the overall metric in the base-to-new generalization setting.

\end{itemize}

\begin{figure*}[h]

	\centering
	\includegraphics[width=0.96\linewidth]{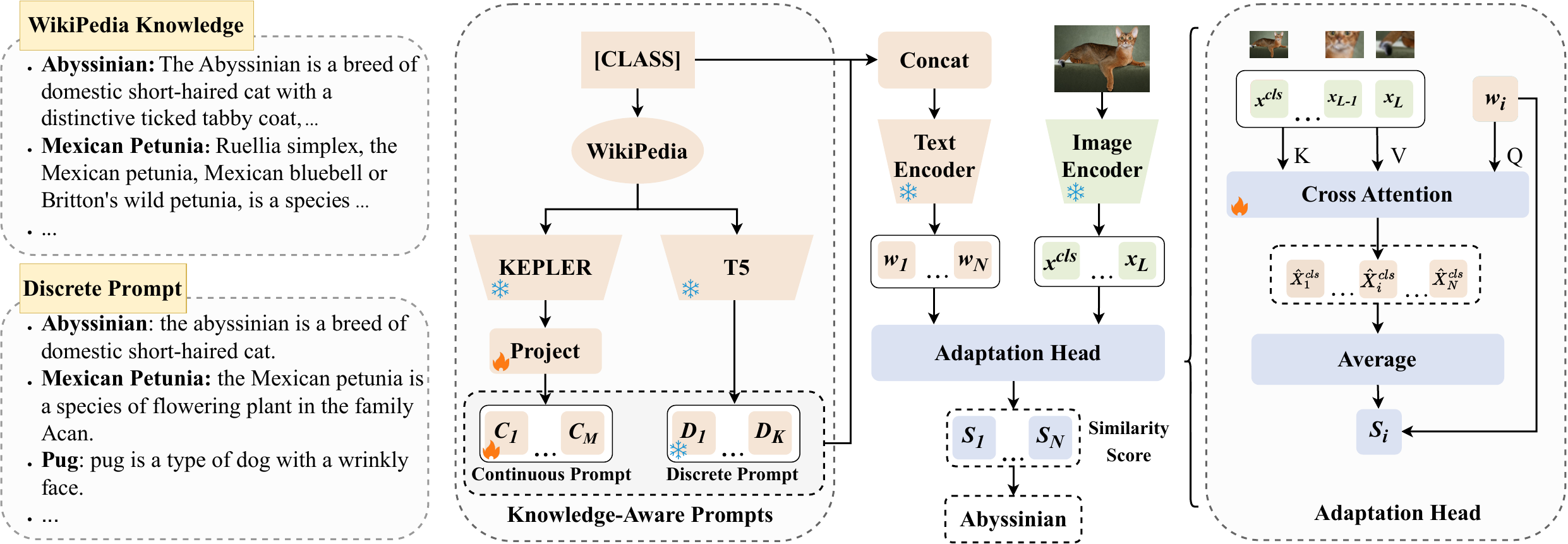}
	\caption{\textbf{Overview of our proposed KAPT (Knowledge-Aware Prompt Tuning) for vision-language models.} We first retrieve textual descriptions of task labels from the external knowledge base. Then, we  construct the discrete and continuous knowledge  prompts to enhance the discrimination of prompts for each category. The discrete, continuous prompts and the class label are concatenated as the input for the text encoder. The output tokens for the image encoder are modulated by an adaptation head, which aggregates task-related visual cues conditioned on high-level text features. The final classification confidence is obtained by calculating the cosine similarity between visual and text embeddings. Wikipedia knowledge (top left) is sourced from Wikipedia Encyclopedia, denoted as \{category:description\}.
 Discrete prompts (lower left) are summaries of corresponding descriptions in Wikipedia. \protect\includegraphics[scale=0.06]{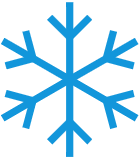} and \protect\includegraphics[scale=0.06]{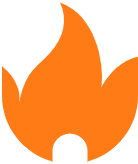} represent frozen and tunable weights during tuning, respectively.}
 \vspace{-0.3cm}
\label{fig3}

\end{figure*}

\section{Related Work}

\subsection{Vision-Language Pre-training}

In recent times, the introduction of extensive image-text data into pre-trained vision-language models has emerged as a prominent trend~\cite{radford2021learning,jia2021scaling,yao2021filip,li2019visualbert,li2020oscar,chen2020uniter,wang2022vlmixer}. A representative work is CLIP \cite{radford2021learning}, which aggregates 400 million image-text pairs from websites, facilitating the vision-language representation learning using a contrastive objective. Contemporary work with CLIP, ALIGN \cite{jia2021scaling} also takes advantage of a large-scale dataset, 1.8 billion noisy image-text pairs, to pre-train a model with contrastive loss. These vision-language models are dual-encoder architectures consisting of an image encoder and a text encoder. Leveraging extensive image-text pairs and dual-encoder architectures, these approaches showcase remarkable prompt-based zero-shot performance across diverse visual classification tasks, by exploiting alignments between text and image features.

\subsection{Prompt Learning}
With the continuous parameter scaling of pre-training models like GPT-3 \cite{brown2020language} and CLIP, fine-tuning \textcolor{black}{the entire} models for downstream tasks becomes daunting because of inefficiency in parameters and potential catastrophic forgetting~\cite{pfeiffer2020adapterfusion}. Notably, recent works \cite{li2021prefix, chen2022knowprompt, zhong2021factual}  have introduced prompt learning that exclusively fine-tune a limited parameter subset, yielding robust results in NLP tasks. Inspired by the swift proliferation of prompt learning within NLP, the computer vision domain has also delved into prompt tuning for resolving downstream tasks~\cite{wang2021actionclip,yao2021cpt,bar2022visual,jia2022visual,wu2023pi,fei2023transferable}. In this context, manual prompt templates within CLIP (\eg, ``a photo of a [\textit{label}]") have been employed for image recognition. However, research in NLP \cite{gao2021making} reveals that identifying suitable manual prompts demands both domain expertise and laborious prompt engineering. Some works \cite{zhou2022learning, rao2022denseclip} adopt learnable continuous prompts to make the vision-language model recall the task-relevant knowledge. Due to the weak generalizability of simple learnable prompts, CoCoOp \cite{zhou2022conditional} proposes conditional prompts by further learning a lightweight neural network to generate an input-conditional token for each image. 
\textcolor{black}{Different from previous works, we advance the generalization of prompt learning by incorporating external knowledge, achieved by integrating accurate descriptions of concepts and their relationships.}

\subsection{External Knowledge Bases}
In natural language processing (NLP), external knowledge bases, such as WordNet \cite{miller1995wordnet} and ConceptNet\cite{speer2017conceptnet}, are frequently harnessed to enhance performance~\cite{ma2021knowledge,yasunaga2021qa}. Early attempts~\cite{wu2022multi, marino2021krisp} in the computer vision community have also verified its effectiveness in visual question answering. Another notable work K-LITE\cite{shen2022k}, distinctively employs external knowledge in the vision-language pre-training phase, yielding visual models with better transferability and sample efficiency. However, within the scope of prompt tuning for vision-language models, there is little attention on harnessing external knowledge for model generalization. Moreover, our knowledge base and task-related knowledge extraction methods, derived from visual entity understanding, distinctly set our work apart from previous NLP research.

\section{Methodology}
\label{sec:meth}
 KAPT is a prompt tuning method that incorporates category-related external knowledge into vision-language pre-training models. Our method builds upon CLIP~\cite{radford2021learning} to effectively leverage its strong zero/few-shot transferability (Section \ref{sec:Pre}). To construct knowledge-aware prompt tuning, we propose two variants of prompts to take full advantage of category-related external knowledge and a task-aware visual adaptation head to adapt the visual representation toward a specific task (Section \ref{sec:KAP}). The overall framework is illustrated in Figure \ref{fig3}.

\subsection{Preliminary: Prompting for CLIP}
\label{sec:Pre}
The CLIP model~\cite{radford2021learning} is a typical dual-encoder architecture consisting of an image encoder and a text encoder. For each image-text pair, an image and the paired text are transformed into high-dimensional  embeddings by image encoder (\eg, ResNet \cite{he2016deep}) and text encoder (\eg, Transformers~\cite{vaswani2017attention}), respectively. The training objective is to align the uni-modal embeddings by contrastive learning, where the model pulls  paired image-text together and pushes the unpaired ones away in latent space. By pre-training on 400M large-scale web datasets, the learned visual representations are discriminative and transferable to zero/few-shot downstream tasks. 

For the zero-shot transfer to image classification with $N$ classes, CLIP constructs a simple prompt ``\textit{a photo of a [\textit{label}]}", 
{{fills \textit{label} with each class name $C_i$, and hence obtains the corresponding textual embedding ${{\mathrm{\mathbf{w}}}_{i}}$ by the text encoder.}} Meanwhile, the image $I$ is fed into the image encoder to generate image embedding ${\rm{\mathbf{x}}}$. The prediction probability is acquired by calculating the cosine similarity between the image embedding and $N$ text embeddings:

\begin{align}\label{eq:1}
p\left( {y = {i}|{\mathrm{\mathbf{x}}}} \right) = \frac{{\exp ({\rm{sim}}({\rm{\mathbf{x}}},{{\rm{\mathbf{w}}}_i})/\tau )}}{{\sum\nolimits_{j = 1}^N {\exp ({\rm{sim}}({\mathrm{\mathbf{x}}},{{\mathrm{\mathbf{w}}}_j})/\tau )} }},
\end{align}
where ${\rm{sim}}( \cdot,\cdot )$ represents the computation of cosine similarity, and $\tau$ represents the temperature ratio.

\subsection{Knowledge-Aware Prompt Tuning (KAPT)}
\label{sec:KAP}
The manually designed prompts of CLIP show inferior performance compared with learnable prompts trained on few-shot datasets. However, learnable prompts usually overfit to seen classes, suffering from the weak generalizability problem in unseen classes. 
To improve the generalizability to novel concepts under the same task, we present a novel prompt tuning method called Knowledge-Aware Prompt Tuning (KAPT). Specifically, we design two complementary knowledge-aware prompts to take full advantage of category-related external knowledge, and we propose a task-aware visual adaptation head to capture task-related visual cues.

\paragraph{Summarized Knowledge.} 
Category-related knowledge is retrieved from the open-source external knowledge base. Note that from existing external knowledge bases \cite{miller1995wordnet, speer2017conceptnet}, it is difficult to retrieve related knowledge for rare or fine-grained concepts, thus resulting in a lower coverage of categories in mainstream vision datasets.
For a high coverage rate of object categories of downstream datasets, we use textual descriptions sourced from Wikipedia Encyclopedia to form a category knowledge base  ${\mathcal{K}}$ for a specific downstream task. However, the resultant textual knowledge is usually expatiatory and contains irrelevant information for visual recognition. To remove redundant information from category-related external knowledge, we feed the knowledge into T5 \cite{raffel2020exploring}, an end-to-end pre-training model which takes text as input and is expected to output modified text summarizing the description of each category. The short text summary is considered as the automatic discrete prompt $\mathrm{D}$ for capturing the key description of categories.

\paragraph{Contextual Knowledge.}
In simplified text descriptions, some category-related information is inevitably filtered as redundant information. In order to make prompts carry contextual information that may cover a broader background of the category, we feed category-related external knowledge retrieved from ${\mathcal{K}}$ into a pre-trained KEPLER~\cite{wang2021kepler} model to generate continuous features. Furthermore, we map obtained continuous features to the multi-modal embedding space by employing a lightweight projector {{which is composed of two linear layers with a bottleneck structure}}. The features are then combined with the context vectors to construct the learnable continuous prompt $\mathrm{C}$ for the corresponding category.  Note that the context vectors are learnable parameters with random initialization.
\begin{table*}[h]
    \tabstyle{6pt}
    \caption{\textbf{Evaluation on the base-to-new generalization setting.} Prompt-based methods learn their prompts from the base classes with 16 shots. 
    We report the accuracy on base and new classes.  
    H (Harmonic mean) evaluates overall performance.}
    \label{tab1}
    \begin{subtable}[t]{.3\textwidth}
    \centering
    \caption{\textbf{Average over 11 datasets}.}
    \begin{tabular}{l cc|c}
    \toprule
    & Base & New & H \\
    \midrule
    CLIP & 64.83   & 70.27 & 67.44 \\
    \midrule
    CoOp & \textbf{79.88} & 59.39  &68.12  \\
    CoCoOp & 76.70  &67.30    & 71.69 \\
    \rowcolor{tabhighlight}
    KAPT & 78.41  & \textbf{70.52} & \textbf{74.26} \\
    \bottomrule
    \end{tabular}
    \end{subtable}
    \vspace{1em}
    \begin{subtable}[t]{.3\textwidth}
    \centering
    \caption{ImageNet.}
    \begin{tabular}{l cc|c}
    \toprule
    & Base & New & H \\
    \midrule
    CLIP & 67.50 & 64.00 & 65.70 \\
    \midrule
    CoOp & \textbf{71.30} & 62.40 & 66.55\\
    CoCoOp & 70.90  & \textbf{66.66} & \textbf{68.71}\\
    \rowcolor{tabhighlight}
    KAPT & 71.10  & 65.20 & 68.02 \\
    \bottomrule
    \end{tabular}
    \end{subtable}
    ~
    \begin{subtable}[t]{.3\textwidth}
    \centering
    \caption{Caltech101.}
    \begin{tabular}{l cc|c}
    \toprule
    & Base & New & H \\
    \midrule
    CLIP & 93.70 & 94.00 & 93.84 \\
    \midrule
    CoOp & \textbf{97.30} & 89.10 & 93.01\\
    CoCoOp & 97.13 & 93.06 & 95.05\\
    \rowcolor{tabhighlight}
    KAPT & 97.10 & \textbf{93.53} & \textbf{95.28} \\
    \bottomrule
    \end{tabular}
    \end{subtable}
    ~
    \begin{subtable}[t]{.3\textwidth}
    \centering
    \caption{OxfordPets.}
    \begin{tabular}{l cc|c}
    \toprule
    & Base & New & H \\
    \midrule
    CLIP & 86.80 & 96.30 & 91.30 \\
    \midrule
    CoOp & 91.20 & 89.73 & 90.46\\
    CoCoOp & \textbf{93.43} & 94.66 & 94.04\\
    \rowcolor{tabhighlight}
    KAPT & 93.13 & \textbf{96.53} & \textbf{94.80} \\
    \bottomrule
    \end{tabular}
    \end{subtable}
    \vspace{1em}
    \begin{subtable}[t]{.3\textwidth}
    \centering
    \caption{StanfordCars.}
    \begin{tabular}{l cc|c}
    \toprule
    & Base & New & H \\
    \midrule
    CLIP & 60.90 & 69.90 & 65.09 \\
    \midrule
    CoOp & \textbf{75.16} & 50.63 & 60.50 \\
    CoCoOp & 65.86 & 66.03 & 65.94\\
    \rowcolor{tabhighlight}
    KAPT & 69.47 & \textbf{66.20} & \textbf{67.79} \\
    \bottomrule
    \end{tabular}
    \end{subtable}
    ~
    \begin{subtable}[t]{.3\textwidth}
    \centering
    \caption{Flowers102.}
    \begin{tabular}{l cc|c}
    \toprule
    & Base & New & H \\
    \midrule
    CLIP & 68.70 & 72.40 & 70.50 \\
    \midrule
    CoOp & \textbf{96.46} & 54.23 & 69.43\\
    CoCoOp & 90.43 & 63.66 & 74.72\\
    \rowcolor{tabhighlight}
    KAPT & 95.00 & \textbf{71.20} & \textbf{81.40} \\
    \bottomrule
    \end{tabular}
    \end{subtable}
    ~
    \begin{subtable}[t]{.3\textwidth}
    \centering
    \caption{Food101.}
    \begin{tabular}{l cc|c}
    \toprule
    & Base & New & H \\
    \midrule
    CLIP & 84.50 & 84.10 & 84.29 \\
    \midrule
    CoOp & 82.80 & 80.30 & 81.53\\
    CoCoOp & \textbf{86.20} & 87.00 & \textbf{86.59}\\
    \rowcolor{tabhighlight}
    KAPT & 86.13 & \textbf{87.06} & \textbf{86.59} \\
    \bottomrule
    \end{tabular}
    \end{subtable}
    \vspace{1em}
    \begin{subtable}[t]{.3\textwidth}
    \centering
    \caption{FGVCAircraft.}
    \begin{tabular}{l cc|c}
    \toprule
    & Base & New & H \\
    \midrule
    CLIP & 20.10 & 28.10 & 23.43\\
    \midrule
    CoOp & \textbf{34.10} & 18.96 & 24.37\\
    CoCoOp & 27.13 & 25.00 & 26.02\\
    \rowcolor{tabhighlight}
    KAPT & 29.67 & \textbf{28.73} & \textbf{29.19} \\
    \bottomrule
    \end{tabular}
    \end{subtable}
    ~
    \begin{subtable}[t]{.3\textwidth}
    \centering
    \caption{SUN397.}
    \begin{tabular}{l cc|c}
    \toprule
    & Base & New & H \\
    \midrule
    CLIP & 69.80 & 73.10 & 71.41 \\
    \midrule
    CoOp & 78.40 & 62.20 & 69.36\\
    CoCoOp & 77.23 &\textbf{74.73} & 75.96\\
    \rowcolor{tabhighlight}
    KAPT & \textbf{79.40} & 74.33 & \textbf{76.78} \\
    \bottomrule
    \end{tabular}
    \end{subtable}
    ~
    \begin{subtable}[t]{.3\textwidth}
    \centering
    \caption{DTD.}
    \begin{tabular}{l cc|c}
    \toprule
    & Base & New & H \\
    \midrule
    CLIP & 53.90 & 58.10 & 55.92 \\
    \midrule
    CoOp & \textbf{77.13} & 41.43 & 53.90\\
    CoCoOp & 73.56 & 52.70 & 61.40\\
    \rowcolor{tabhighlight}
    KAPT & 75.97 & \textbf{58.30} & \textbf{65.97} \\
    \bottomrule
    \end{tabular}
    \end{subtable}
    ~
    \begin{subtable}[t]{.3\textwidth}
    \centering
    \caption{EuroSAT.}
    \begin{tabular}{l cc|c}
    \toprule
    & Base & New & H \\
    \midrule
    CLIP & 43.40 & 61.40 & 50.85 \\
    \midrule
    CoOp & \textbf{92.13} & 51.83 & 66.34\\
    CoCoOp & 82.33 & 50.00 & 62.21\\
    \rowcolor{tabhighlight}
    KAPT & 84.80 & \textbf{67.57} & \textbf{75.21} \\
    \bottomrule
    \end{tabular}
    \end{subtable}
    ~
    \begin{subtable}[t]{.3\textwidth}
    \centering
    \caption{UCF101.}
    \begin{tabular}{l cc|c}
    \toprule
    & Base & New & H \\
    \midrule
    CLIP &  63.90&  71.60 & 67.53  \\
    \midrule
    CoOp & \textbf{82.76} &52.53  &64.27 \\
    CoCoOp & 79.50 & 66.76  &72.57 \\
    \rowcolor{tabhighlight}
    KAPT &80.83  & \textbf{67.10} & \textbf{73.33} \\
    \bottomrule
    \end{tabular}
    \end{subtable}
  \vspace{-0.2cm}
\end{table*}

\paragraph{Knowledge-Aware Prompts.}
To take full advantage of category-related external knowledge and considering learning-based continuous prompts have a higher risk of overfitting towards seen classes than discrete prompts, we concatenate learnable continuous prompt $\mathrm{C{_{{y_i}}}}$, ${y_i}$, and automatic discrete prompt $\mathrm{D{_{{y_i}}}}$ to build knowledge-aware prompt for category ${y_i}$. Feeding knowledge-aware prompts into the text encoder ${f^T}\left(  \cdot  \right)$, we obtain the text features ${\mathrm{\mathbf{W}}} = \{ {{\mathrm{\mathbf{w}}}_i}\} _{i = 1}^N$ which are used to calculate similarity with visual representations and  provided to adaptation head as the salient cues relevant to categories.

\paragraph{Adaptation Head.}
{{Given an image $I$, the image encoder  ${f^I}\left(  \cdot  \right)$ transforms $I$ into a set of visual feature vectors }} ${\mathrm{\mathbf{X}}} = [{{\mathrm{\mathbf{x}}}^{cls}},{{\mathrm{\mathbf{x}}}_1}, \ldots ,{{\mathrm{\mathbf{x}}}_j}, \ldots ,{{\mathrm{\mathbf{x}}}_L}]$. To further adapt the visual representation towards a specific task to reduce disturbance of task-irrelated visual concepts, we construct the task-aware visual adaptation head to focus the attentive visual features by attending to the salient cues relevant to categories. For each text features ${{\mathrm{\mathbf{w}}}_i}$, we take it as query vector, and image features ${\mathrm{\mathbf{X}}}$ as the key and value vector. 
\begin{align}\label{eq:4}
\hat {\mathrm{\mathbf{X}}}_i^{cls} = LN({{\mathrm{\mathbf{x}}}^{cls}} + {\mathrm{CrossAttention}}({{\mathrm{\mathbf{w}}}_i};{\mathrm{\mathbf{X}}})),
\end{align}
where $\hat {\mathrm{\mathbf{X}}}_i^{cls}$ represents the enhanced image features, ${\mathrm{CrossAttention}( \cdot )}$ refers to  cross attention, $LN$ is Layer Normalization and ${x^{cls}}$ denotes the ${\mathrm{\mathbf{[CLS]}}}$ token of the input image. To converge all information of $\hat{\mathrm{\mathbf{X}}} _i^{cls}$, we obtain the mean image features by computing the average of all enhanced image features $[\hat {\mathrm{\mathbf{X}}}_i^{cls}]_{i = 1}^N$.
\begin{align}\label{eq:5}
{{\bar {\mathrm{\mathbf{X}}}}} = {\mathrm{AvgPool}}([\hat {\mathrm{\mathbf{X}}}_i^{cls}]_{i = 1}^N)=\frac{1}{N}\sum\limits_{i = 1}^N {\hat X_i^{cls}},
\end{align}
where ${{\bar {\mathrm{\mathbf{X}}}}}$ represents average-pooled image features. We then compute the cosine similarity between ${{\bar {\mathrm{\mathbf{X}}}}}$ and text features ${\mathrm{\mathbf{W}}}$,
\begin{align}\label{eq:6}
p\left( {y = {{i}}|{{\bar {\mathrm{\mathbf{X}}}}}} \right) = \frac{{\exp (\rm{sim}({{\bar {\mathrm{\mathbf{X}}}}},{{\mathrm{\mathbf{w}}}_i})/\tau )}}{{\sum\nolimits_{j = 1}^N {\exp (\rm{sim}({{\bar {\mathrm{\mathbf{X}}}}},{{\mathrm{\mathbf{w}}}_j})/\tau )} }},
\end{align}
where $\tau$ is a learned temperature parameter and ${\mathrm{sim}( \cdot )}$ denotes cosine similarity.

\paragraph{Training Objective.}
During training, we update the gradients of knowledge-aware prompts and adaptation head while keeping the parameters of CLIP frozen. 
The training objective is to minimize he cross-entropy loss:
\begin{align}\label{eq:7}
 {\mathcal{L}_{ce}}=  - \sum\limits_i {{}\log p\left( {y = {{i}}|{{\bar {\mathrm{\mathbf{X}}}}}} \right),1 \le i \le N}.
\end{align}

\begin{table*}[t]
\caption{\textbf{Model comparison in the one-shot setting.} We fine-tune KAPT and other models mentioned above on 11 datasets with only one sample in each category, where KAPT performs best in the average.}
\vspace{-.5em}

\small
\subfloat[
\textbf{Comparison to CoOp and CoCoOp in one-shot setting.}
    \label{tab:2}
]{

\begin{minipage}{0.75\linewidth}{\begin{center}
\renewcommand\arraystretch{1.0}

        \setlength{\tabcolsep}{0.55 mm}{
            \begin{tabular}{lcccccccccccc}
            \toprule
            Method & \begin{tabular}[c]{@{}c@{}}\small Image\\ Net\end{tabular} & \begin{tabular}[c]{@{}c@{}}\small Caltech\\ 101\end{tabular} & \begin{tabular}[c]{@{}c@{}}\small Oxford\\ Pets\end{tabular} & \begin{tabular}[c]{@{}c@{}}\small Stanford\\ Cars\end{tabular} & \begin{tabular}[c]{@{}c@{}}\small Flowers\\ 102\end{tabular} & \begin{tabular}[c]{@{}c@{}}\small Food\\ 101\end{tabular} & \begin{tabular}[c]{@{}c@{}}\small FGVC\\ Aircraft\end{tabular} & \begin{tabular}[c]{@{}c@{}}\small SUN\\ 397\end{tabular} & \small DTD            & \begin{tabular}[c]{@{}c@{}}\small Euro\\ SAT\end{tabular} & \begin{tabular}[c]{@{}c@{}}\small UCF\\ 101\end{tabular} & \small Avg. \\  \midrule
            CoOp                &      60.20    & 91.16          & 86.43          & 60.56          & 71.06          & 74.73          & 15.23          & 63.90          & 47.66          & \textbf{55.46}          & 66.23          &     62.96    \\\noalign{\smallskip}
            CoCoOp               & \textbf{64.50}    & \textbf{92.43} & 87.43          & 60.60          & 68.30          & \textbf{79.46} & 10.26          & \textbf{65.66} & 45.46          & 48.60& \textbf{67.56} & 62.75   \\\noalign{\smallskip}
            KAPT              &     62.90        & 89.63          & \textbf{87.60} & \textbf{60.73} & \textbf{74.17} & 78.07          & \textbf{22.13} & 64.50          & \textbf{50.93} & 46.50          & 65.90          &  \textbf{63.91}       \\ \bottomrule
            \end{tabular}
            }
    \label{tab:auxLoss}
\end{center}}\end{minipage}
}
\hspace{0.5em}
\subfloat[
\textbf{Model ablations.}
    \label{tab:abla}
]{

\begin{minipage}{0.20\linewidth}{\begin{center}
\renewcommand\arraystretch{1.11}

        \setlength{\tabcolsep}{1.0 mm}{
        \begin{tabular}{l |c}
            \toprule
            Method &Avg. \\
            \midrule
            Baseline & 62.96 \\
            w/ knowledge & 63.47 \\
            w/ adaptation  & 63.20\\
            KAPT & \textbf{63.91} \\         
            \bottomrule
        \end{tabular}
        }
\label{tab:auxLoss}
\end{center}}\end{minipage}
}
\vspace{-0.5cm}
\end{table*}

\section{Experiments}
\subsection{Experimental Setup}
\label{sec:Es}
\paragraph{Datasets.}
For evaluation, we perform extensive experiments on 11 image classification datasets: Flowers102~\cite{nilsback2008automated}, OxfordPets~\cite{parkhi2012cats},  Food101~\cite{bossard2014food}, StanfordCars~\cite{krause20133d}, FGVCAircraft~\cite{maji2013fine}, SUN397~\cite{xiao2010sun}, DTD~\cite{cimpoi2014describing}, EuroSAT~\cite{helber2019eurosat}, UCF101~\cite{soomro2012ucf101}, Caltech101~\cite{fei2004learning}, and ImageNet~\cite{deng2009imagenet}. These datasets cover a variety of fine-grained classification tasks, building an all-around
benchmark, including species of plants or animals, satellite imagery of traffic, and diverse general objects. Meanwhile, we use Wikidata5m \cite{wang2021kepler}, the main source of external knowledge, which is a large-scale knowledge graph dataset with aligned text descriptions from the corresponding Wikipedia pages.

\paragraph{Training Details.} We adopt ViT-B/32 as the backbone network for all experiments. For constructing automatic discrete prompts, we set the maximum length of discrete prompts and beam search as 20 and 6, respectively. We fix the number of context tokens for automatic continuous prompts to 16. The adaptation head is configured with the dropout rate of 0.1 and 8 attention heads. Throughout training, the initial learning rates for both the knowledge-aware prompts and adaptation head are established at 0.002 and 0.005, respectively, with adjustment governed by the cosine annealing rule. Our optimization approach employs SGD, and we set the maximum epoch count to 50. To ensure fair comparisons with previous works, we average the three scores with different seeds. All experiments are conducted on a single Tesla V100 GPU with 32GB memory, within the PyTorch framework. 

\paragraph{{{Compared Methods.}}} We compare KAPT with existing representative prompting methods based on CLIP. CoOp \cite{zhou2022learning} is the landmark prompt tuning method, which learns the context prompt to make CLIP recall the task-relevant knowledge for downstream image recognition. To overcome the deficiency of learnable prompts on generalization ability, CoCoOp \cite{zhou2022conditional} proposes conditional prompts with a network module to improve the generalization on unseen classes. Note that the results of baseline models are obtained by the released official codes.

\begin{table*}[t]
\caption{{\textbf{Robustness evaluation to distribution shift.} {\color{black}All models are trained with 16-shot samples.}}
} \label{tab:robustness}
\vspace{-0.5em}
\small
\subfloat[
\textbf{Cross-dataset transfer (source: ImageNet or SUN397).}
    \label{tab:cross-dataset}
]{

\begin{minipage}{0.7\linewidth}{\begin{center}
\renewcommand\arraystretch{1.0}
\setlength{\tabcolsep}{1mm}{
\begin{tabular}{lcccccccccc}
\toprule
    Method & \begin{tabular}[c]{@{}c@{}}\small Caltech\\ 101\end{tabular} & \begin{tabular}[c]{@{}c@{}}\small Oxford\\ Pets\end{tabular} & \begin{tabular}[c]{@{}c@{}}\small Stanford\\ Cars\end{tabular} & \begin{tabular}[c]{@{}c@{}}\small Flowers\\ 102\end{tabular} & \begin{tabular}[c]{@{}c@{}}\small Food\\ 101\end{tabular} & \begin{tabular}[c]{@{}c@{}}\small FGVC\\ Aircraft\end{tabular}  & \small DTD            & \begin{tabular}[c]{@{}c@{}}\small Euro\\ SAT\end{tabular} & \begin{tabular}[c]{@{}c@{}}\small UCF\\ 101\end{tabular} & \small Avg.  \\ 
\midrule
\multicolumn{5}{l}{{\textit{Source dataset: ImageNet}}} \\
CoCoOP      &\textbf{92.15} & 88.90 & \textbf{60.30} & 65.80 &  \textbf{80.65} & 17.50  & 40.05 & \textbf{41.70} & 64.20 & 61.25    \\
KAPT        & 88.90 & \textbf{89.40} & 58.15 & \textbf{68.00} & 79.95 & \textbf{17.95}  & \textbf{44.80} & 41.35 & \textbf{65.05}  & \textbf{61.50} \\ 
\midrule
\multicolumn{5}{l}{{\textit{Source dataset: SUN397}}} \\
CoCoOP      &\textbf{90.95} & 80.20  & 52.00 & 57.70 &  76.70 & 12.15 &37.90 & \textbf{43.60}  & \textbf{66.35} & 57.50 \\
KAPT        &90.20 & \textbf{85.45} & \textbf{53.85} & \textbf{61.80} & \textbf{78.10} & \textbf{14.55}  & \textbf{42.95} & 33.30 & 59.90 & \textbf{57.78}    \\ 

\bottomrule
\end{tabular}}

\end{center}} \end{minipage}
}
\hspace{0.5em}
\subfloat[
\textbf{Domain generalization.}
    \label{tab:domain-generalization}
]{
\small
\begin{minipage}{0.3\linewidth}{\begin{center}
\renewcommand\arraystretch{1.33}
        \setlength{\tabcolsep}{2 mm}{
        \begin{tabular}{c|cc}
\toprule
\multicolumn{3}{c}{{\textit{Source dataset: ImageNet}}} \\
\midrule
Target    & CoCoOp & KAPT  \\ 
\midrule
INV2      & \textbf{58.40}  & 58.10 \\
IN-S      & 42.00  & \textbf{42.30} \\
IN-A      & \textbf{31.60}  & 31.10 \\
IN-R      & 66.30  & \textbf{66.60} \\ \bottomrule
\end{tabular}
        }

\end{center}}\end{minipage}
}
\vspace{-0.2em}
\end{table*}

\subsection{Comparison with State-of-the-Art Methods}
\label{sec:CwS}
The performance of KAPT and two baselines are shown in Table \ref{tab1}. KAPT achieves outstanding performance and establishes state-of-the-art results on the overall accuracy (evaluated by harmonic mean). Compared to CoOp, KAPT demonstrates an overall performance improvement across all datasets, with a notable increase of 11.13\% particularly on unseen classes. As the generalization to unseen classes is an essential capability of models, CoCoOp proposes conditional prompts to improve the generalizability. Although CoCoOp achieves a better performance than CoOp, KAPT still makes considerable progress compared with CoCoOp in average scores of all metrics. When contrasting KAPT with CoCoOp, improvements of 1.71\%, 3.22\%, and 2.57\% are observed in base classes, new classes, and harmonic mean, respectively. Importantly, KAPT outperforms CoCoOp across 10 out of 11 datasets. Zhou et al. \cite{zhou2022conditional} claim that CLIP is a strong competitor in unseen classes due to learning-based prompts easily overfitting to base classes than manual prompts. Compared with zero-shot CLIP, KAPT achieves an absolute gain of 0.25\% on new classes and outperforms CLIP on 7 out of 11 datasets, including ImageNet, DTD, EuroSAT, OxfordPets, Food101, SUN397, and AirCraft101. However, it's worth noting that CoCoOp's accuracy on new classes outperforms zero-Shot CLIP only on ImageNet and SUN397.

 \subsection{One-shot Classification Performance}
KAPT performs excellently on the generalization test, demonstrating outstanding overall performance on the base-to-new setting. Meantime, the anti-overfitting ability is also essential for KAPT. Here, we train KAPT and other baseline methods in the one-shot setting to evaluate anti-overfitting ability. Table \ref{tab:2} shows the comparisons of KAPT with other models on 11 datasets. Overall, our KAPT shows its superiority over baseline models on average performance on one-shot settings. It is well known that CoOp trained on closed datasets has strong performance on seen classes in closed datasets. However, our KAPT still beats CoOp on 8 out of 11 datasets under closed datasets and one-shot setting.  In Table \ref{tab:abla}, the integration of the knowledge-aware prompt and the adaptation head has demonstrated remarkable achievement. The success of this approach can be attributed to the fact that when the downstream task has limited samples, category-related external knowledge is able to make up the lack of visual information to a certain extent and assist the adaptation head in filtering out irrelevant information, which is not associated with the category.

\begin{table}[]
    \centering
    \small
    \caption{\textbf{Ablation  study  over KAPT components.} Average accuracy (\%) over 11 datasets is reported. }\label{tab3}
    \begin{tabular}{l cc|c}
    \toprule
    Method & Base & New & H \\\midrule
    Baseline  & 79.41 & 64.02 & 70.88 \\
    w/ knowledge & 77.70  & 67.24 & 72.09  \\
    w/ adaptation head & \textbf{80.09} & 65.58 & 72.11  \\
    KAPT & 78.41  & \textbf{70.52} & \textbf{74.26}\\
    \bottomrule
    \end{tabular}
\end{table}

\subsection{Distribution Shift Robustness}

\textcolor{black}{We systematically evaluated the robustness of KAPT under distribution shifts, \ie, cross-dataset transfer and domain generalization scenarios. In the cross-dataset transfer scenario, we train two models seperately using ImageNet and SUN397 datasets. Subsequently, our approach's performance was assessed across 9 diverse datasets. As evident in Table \ref{tab:cross-dataset}, when utilizing ImageNet as the source dataset, KAPT exhibits a modest improvement in transferability compared to CoCoOp, achieving an average accuracy of 61.50\%, surpassing CoCoOp by 0.25\%. Similarly, when SUN397 is employed as the source dataset, KAPT showcases slightly better performance relative to CoCoOp, outperforming CoCoOp across 6 out of 9 datasets and yielding an average accuracy gain of 0.28\% over CoCoOp. 
}

Furthermore, in order to study the robustness of our method to domain generalization, we evaluate the transferability of the model trained on ImageNet (IN) to various out-of-domain datasets, \ie, ImageNetV2 (INV2) \cite{recht2019imagenet}, ImageNet-Sketch (IN-S) \cite{wang2019learning}, ImageNet-A (IN-A) \cite{hendrycks2021natural} and ImageNet-R (IN-R) \cite{hendrycks2021many}. 
By observing Table \ref{tab:domain-generalization}, it becomes evident that KAPT and CoCoOp yield comparable performance in the domain generalization. We conclude that the proposed knowledge enhances the transferability to new categories, while robust to the
domain change on seen categories.

\begin{table*}[h]
\small
\centering
\caption{ \textbf{Ablation study over types of prompts.} We report the accuracy of KAPT (discrete + continuous) and other variant models (with continuous only or with discrete only) across the 11 datasets in the unseen setting. 
 }\label{tab4}
 \resizebox{\linewidth}{!}{
\begin{tabular}{lcccccccccccc}
\toprule
Method & \begin{tabular}[c]{@{}c@{}}\small Image\\ Net\end{tabular} & \begin{tabular}[c]{@{}c@{}}\small Caltech\\ 101\end{tabular} & \begin{tabular}[c]{@{}c@{}}\small Oxford\\ Pets\end{tabular} & \begin{tabular}[c]{@{}c@{}}\small Stanford\\ Cars\end{tabular} & \begin{tabular}[c]{@{}c@{}}\small Flowers\\ 102\end{tabular} & \begin{tabular}[c]{@{}c@{}}\small Food\\ 101\end{tabular} & \begin{tabular}[c]{@{}c@{}}\small FGVC\\ Aircraft\end{tabular} & \begin{tabular}[c]{@{}c@{}}\small SUN\\ 397\end{tabular} & \small DTD            & \begin{tabular}[c]{@{}c@{}}\small Euro\\ SAT\end{tabular} & \begin{tabular}[c]{@{}c@{}}\small UCF\\ 101\end{tabular} & \small Average \\  \midrule
discrete + continuous                  &     65.20     & \textbf{93.53}          & \textbf{96.53} & 66.20          & \textbf{71.20} &     87.06     & \textbf{28.73} & 74.33          & \textbf{58.30} & \textbf{67.57} & 67.10          &   \textbf{70.52}      \\ \noalign{\smallskip}
w/ continuous only          &      \textbf{66.06}   & 93.46 & 96.10          & \textbf{66.96} & 66.63          & \textbf{87.46} & 11.30          & \textbf{74.66} & 51.76          & 58.23          & \textbf{68.73} &       67.53  \\ \noalign{\smallskip}
w/ discrete only         &     64.36     & 92.93          & 96.33          & 63.80          & 66.93          & 85.00          & 23.70          & 71.10          & 57.86          & 64.60          & 66.33          &  68.86       \\ \bottomrule
\end{tabular}
}
\vspace{-0.2cm}
\end{table*}

\subsection{Ablation Study}
We investigate the importance of the critical components of KAPT for its excellent generalization ability through a series of ablation experiments. Initially, we assess the performances of two variants, namely, ablated knowledge-aware prompt or adaptation head, in base-to-new settings to ascertain their necessity. Subsequently, we quantitatively examine the essential role of discrete and learnable continuous prompts in enhancing KAPT's generalization ability under unseen setting. Lastly, additional experiments offer further hyperparameter analysis.

\paragraph{Effectiveness of Proposed Components.}
\label{sec:EDC}

In our framework, knowledge-aware prompts and adaptation head are two core components. To investigate the effectiveness of each component, we conduct  ablation experiments to reveal how the combination of two modules improves the overall performance of KAPT, especially on unseen classes. Specifically, we carry out the experiments by adding them one by one to observe the change in overall performance. We take CoOp as the baseline. The experimental results are shown in Table \ref{tab3}. We can find that adding the adaptation head helps to improve the accuracy on base, new and harmonic by a margin of 0.68\%, 1.56\%, and 1.23\% compared with the baseline method. The result demonstrates that the adaptation head not only adopts visual representation towards a specific task but also helps improve the model's generalizability on unseen classes in the same task. {{Moreover, we also find that adding knowledge-aware prompts can improve the baseline 
by 3.22\% and 1.21\% on new accuracy and harmonic mean.}} Although the base accuracy of adding knowledge-aware prompts drops below the baseline model, the gains on unseen classes are far outweighed by the losses on seen classes. The result demonstrates that knowledge-aware prompts are able to relieve the weak generalizability problem greatly. Compared to the baseline model, KAPT constructed by knowledge-aware prompts and adaptation head shows an improvement of 6.50\% and 3.38\% on classification accuracy on unseen classes and harmonic mean.

\begin{figure}[]
	\centering
	\includegraphics[width=1.0\linewidth]{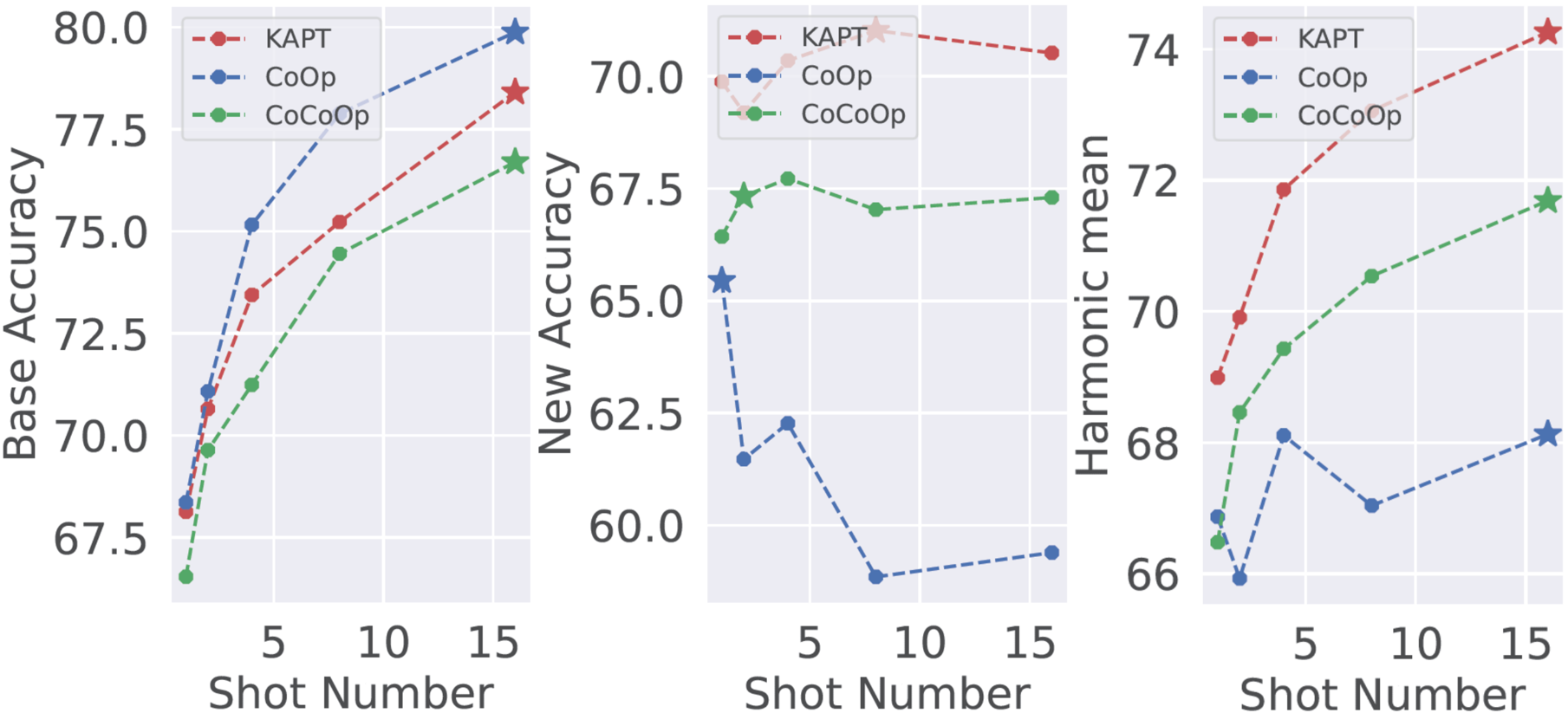}
    \caption{\textbf{Performance comparisons with different shot numbers}. KAPT and CoOp/CoCoOp are trained on 11 benchmark datasets with varying shots on the base-to-new generalization setting.  KAPT outperforms other models from 1-shot to 16-shot settings.}
 \label{fig4}
\vspace{-0.5cm}
\end{figure}

\paragraph{Generalizability of Knowledge Prompts.}
\label{sec:GKP}

The construction of automatic prompts is an essential part of KAPT. Although we analyze the importance of knowledge-aware prompts in Section \ref{sec:EDC}, we still need to explore the independent role and mutual influence of two different automatic prompts. We consider two variant models: i) KAPT (with continuous only) and ii) KAPT (with discrete only). As shown in Table \ref{tab4}, KAPT outperforms two variant models on most datasets and obtains the best average accuracy on unseen classes. This demonstrates that the cooperation of the two kinds of automatic prompts can promote the improvement of the generalization ability of the model. Meanwhile, we discover that the average results of   KAPT (with continuous only) and KAPT (with discrete only)  outperform   KAPT (with adaptation head) in Table \ref{tab3} by 1.95\% and 3.28\%  on unseen classes, respectively. The above findings prove that any automatic prompts sourced from related-category external knowledge help the adaptation head to remove task-irrelevant concepts  to improve its generalization ability. 
However, we also notice that our KAPT underperforms KAPT (with continuous only ) on 5 out of 11 datasets. This is probably because some noise information is still retained in discrete prompts and the qualities of textual descriptions between the different domains in the Wikipedia Encyclopedia have significant differences. Thus, we leave more sophisticated data processing as future work.

\begin{figure}[]
	\centering
	\includegraphics[width=1\linewidth]{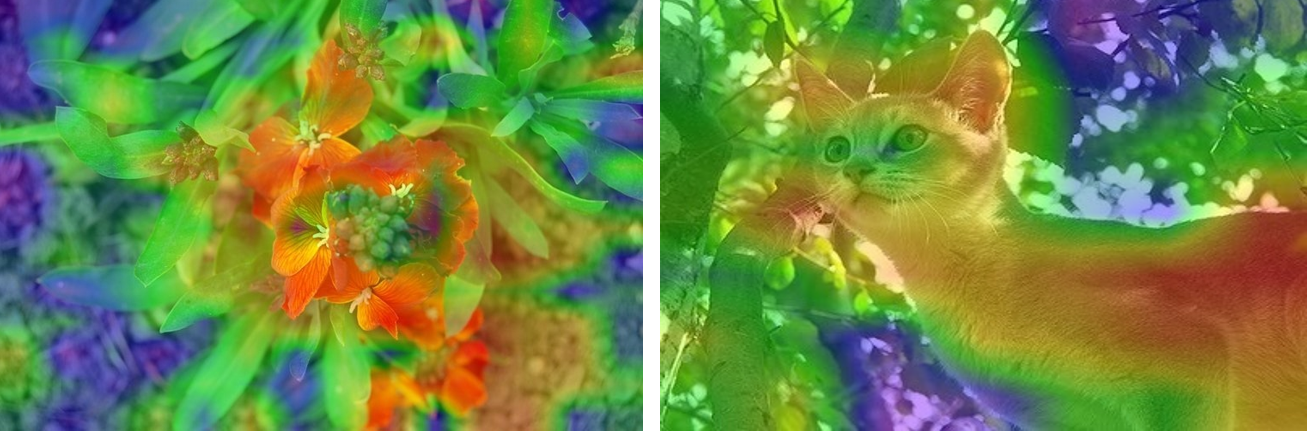}
	\caption{\textbf{Attention heatmaps of adaptation head. }Two images are from Flowers102 and OxfordPets, respectively.} \label{fig6}
\vspace{-1.5em}
\end{figure}

\paragraph{Shot Number.} 
 Section \ref{sec:CwS} reports the performance of KAPT, CoOp, and CoCoOp in 16-shot setting. Here, we want to investigate further the effect of different shot settings on base-to-new generalization setting. Therefore, we conduct experiments with varying shot values on KAPT, \ie, 1, 2, 4, 8, 16. 
 By examining the outcomes depicted in Figure \ref{fig4}, it becomes evident that the performance of both KAPT and CoCoOp progressively improves with the augmentation of shot numbers in the new setting and overall performance. In contrast, the performance of CoOp exhibits instability with increasing sample numbers. This erratic behavior in CoOp's results might arise from its tendency to overfit to seen classes, resulting in diminished accuracy on unseen classes. In contrast, KAPT achieves an optimal overall performance by striking a skillful balance between seen and unseen classes.

\begin{table}[]
\small
\centering
\caption{\textbf{Performance comparisons using different vision backbones from CLIP.} KAPT outperforms CoOp and CoCoOp in three vision backbones overall and even achieves better performance in all settings than CoCoOp. \textcolor[HTML]{006896}{$\Delta$} denotes absolute gains of KAPT over CoCoOp.} \label{tab5}
\setlength{\tabcolsep}{0.3 mm}{
\begin{tabular}{l|ccc|ccc|ccc}
\toprule
                      \multirow{2}{*}{Method} & \multicolumn{3}{c|}{ResNet-50}                                           & \multicolumn{3}{c|}{ViT-B/32}                                                      & \multicolumn{3}{c}{ViT-B/16}                                                      \\ 
\multicolumn{1}{c|}{} & Base                 & New                  & H                     & Base                      & New                       & H                          & Base                      & New                       & H                         \\ \midrule

 CoOp                  & \textbf{77.54}               & 57.48               & 66.02                 & \textbf{79.88}            & 59.39                     & 68.12                      & \textbf{82.83}            & 62.82                    & 71.45                    \\
CoCoOp                & 75.26                & 64.36                & 69.39                 & 76.70                     & 67.30                     & 71.69                     & 79.53                     & 71.49                     & 75.29                     \\ \midrule
KAPT    & { 75.39 }      & { \textbf{64.71}}      & { \textbf{69.94}}                & 78.41                     & \textbf{70.52}            & \textbf{74.26}             & 81.10                     & \textbf{72.24}            & \textbf{76.41}            \\ 
 \multicolumn{1}{l|}{\textcolor[HTML]{006896}{$\Delta$}} & \multicolumn{1}{l}{\textcolor[HTML]{006896}{+0.13}} & \multicolumn{1}{l}{\textcolor[HTML]{006896}{+0.35}} & \multicolumn{1}{l|}{\textcolor[HTML]{006896}{+0.55}} & \multicolumn{1}{l}{\textcolor[HTML]{006896}{+1.71}} & \multicolumn{1}{l}{\textcolor[HTML]{006896}{+3.22}} & \multicolumn{1}{l|}{\textcolor[HTML]{006896}{+2.57}} & \multicolumn{1}{l}{\textcolor[HTML]{006896}{+1.57}} & \multicolumn{1}{l}{\textcolor[HTML]{006896}{+0.75}} & \multicolumn{1}{l}{\textcolor[HTML]{006896}{+1.12}} \\ \bottomrule
\end{tabular}
}
      \vspace{-1.5em}
\end{table}

{{\paragraph{Visualization of the Adaptation Head.} We conducted experiments to analyze the adaptation head's ability to filter out task-irrelevant information. For this analysis, we utilized images from OxfordPets and Flowers102 datasets as examples. As depicted in Figure \ref{fig6}, the model pays more attention to task-related content, like flowers and cats. Therefore, the adaptation head aligns the visual representation towards the specific task, mitigating the interference from task-irrelated visual concepts.}}

\paragraph{Backbone Models.} CLIP provides a variety of vision backbones, such as ResNet-50, ViT-B/32 and ViT-B/16. We assess the performance of KAPT not only with ViT-B/32 as our backbone but also with ResNet-50 and ViT-B/16. Table \ref{tab5} presents the averaged performance across 11 datasets using different backbones. As anticipated, KAPT consistently outperforms both CoOp and CoCoOp across various backbone architectures.

{{\paragraph{Parameter Number.} Considering the introduction of trainable parameters through the adaptation head, we empirically investigated their influence on model performance. To ensure fair evaluation, we devised {CoCoOp$\dag$} to increase the parameter number of CoCoOp to the same level as our models.  As presented in Table \ref{tab6}, the results verify the notable advantages of our method, even when parameters in both approaches are matched in scale.}}

\section{Conclusion}
To overcome potential overfitting towards seen classes and underperforming generalizability in unseen scenarios under the same task, we propose a knowledge-aware prompt tuning (KAPT) for vision-language models. 
Specifically, we identify the importance of exploring category-related external knowledge by designing two types of knowledge-aware prompts for text. Further, we also present an adaptation head to adapt the visual representation toward a specific task.
Extensive experiments validate KAPT's superiority over state-of-the-art approaches on standard benchmark datasets. However, since KAPT builds upon the CLIP backbone, inherent biases and fairness concerns from the original model may persist during prompt learning. While our model exhibits enhanced performance, further refinements are possible. To achieve broader coverage of visual concepts, Wikipedia Encyclopedia and other external knowledge bases could be jointly used as the knowledge source. Meanwhile, existing external knowledge bases are generally diverse in the open domain, often lacking task-specific expertise. Constructing multi-source knowledge bases with specialized expertise remains future investigation.

{\color{black}\paragraph{Acknowledgments.} This work was supported by the National Key R\&D Program of China (Grant NO. 2022YFF1202903) and the National Natural Science Foundation of China (Grant NO. 62122035).}

\begin{table}[]
\small
\centering
    \caption{\textbf{Parameter Comparison.} {{{$\dag$} means CoCoOP with more prompt tokens and larger dimension of Meta-Net.}}}\label{tab6}
\begin{tabular}{ll|c|cc|c}
\toprule
\multicolumn{2}{l|}{Method}        & Parameters & Base  & New   & H     \\ \midrule
\multicolumn{2}{l|}{CoCoOp} & 0.4M      & 76.70& 67.30 & 71.69 \\ 
\multicolumn{2}{l|}{CoCoOp$\dag$} & 1.5M       & 76.79 & 68.62 & 72.47 \\ \midrule
\multicolumn{2}{l|}{KAPT}   & 1.3M       & \textbf{78.41} & \textbf{70.52} & \textbf{74.26} \\ \bottomrule
\end{tabular}
\vspace{-1.5em}
\end{table}

{\small
\bibliographystyle{ieee_fullname}
\bibliography{egbib}
}
\end{document}